\documentclass{article}
\usepackage{ijcai19}

\usepackage{times}
\usepackage{soul}
\usepackage{url}
\usepackage[hidelinks]{hyperref}
\usepackage[utf8]{inputenc}
\usepackage[small]{caption}
\usepackage{graphicx}
\usepackage{amsmath}
\usepackage{booktabs}
\usepackage{algorithm}
\usepackage{algorithmic}
\usepackage{latexsym} 
\usepackage{amsfonts} 
\usepackage{multirow} 
\usepackage{subfigure} 
\usepackage{xspace}
\usepackage{color}
\usepackage{colortbl}
\usepackage[english]{babel}
\usepackage{balance}
\usepackage{amsmath,bm}
\usepackage{subfigure}
\usepackage{float}

\definecolor{Gray}{gray}{0.9}

\title{Relation-Aware Entity Alignment for Heterogeneous Knowledge Graphs}

\author{
Yuting Wu$^1$\and
Xiao Liu$^1$\and
Yansong Feng$^1$\footnote{Corresponding author.}\and
Zheng Wang$^2$\and
Rui Yan$^1$\And
Dongyan Zhao$^1$\\
\affiliations
$^1$Institute of Computer Science and Technology, Peking University, China\\
$^2$School of Computing and Communications, Lancaster University, U. K.\\
\emails
\{wyting, lxlisa, fengyansong, ruiyan, zhaodongyan\}@pku.edu.cn,
z.wang@lancaster.ac.uk 
}

\begin{document}

\maketitle

\begin{abstract}
	
	Entity alignment is the task of linking entities with the same real-world identity from different knowledge graphs (KGs), which has been recently dominated by embedding-based methods. Such approaches work by learning KG representations so that entity alignment can be performed by measuring the similarities between entity embeddings. While promising, prior works in the field often fail to properly capture complex relation information that commonly exists in multi-relational KGs, leaving much room for improvement. In this paper, we propose a novel Relation-aware Dual-Graph Convolutional Network (RDGCN) to incorporate relation information via attentive interactions between the knowledge graph and its dual relation counterpart, and further capture neighboring structures to learn better entity representations. Experiments on three real-world cross-lingual datasets show that our approach delivers better and more robust results over the state-of-the-art alignment methods by learning better KG representations.
	
\end{abstract}

\section{Introduction}
\label{section:intro} 
Knowledge graphs (KGs) are the building blocks for various AI applications like question-answering \cite{zhang2018variational}, text classification \cite{wang2016text}, recommendation
systems \cite{zhang2016collaborative}, etc. Knowledge in KGs is usually organized into triples of $\langle$\emph{head entity, relation, tail entity}$\rangle$. 
There are considerable works on knowledge representation learning to construct distributed representations for both entities and relations. Exemplary works are the so called \emph{trans-family} methods like
TransE \cite{bordes2013translating}, TransH \cite{wang2014knowledge}, and PTransE \cite{lin2015modeling},
which interpret a relation as the translation operating on the embeddings of its head entity and tail entity. 

\begin{figure}
	\centering
	\includegraphics[width=0.7\linewidth]{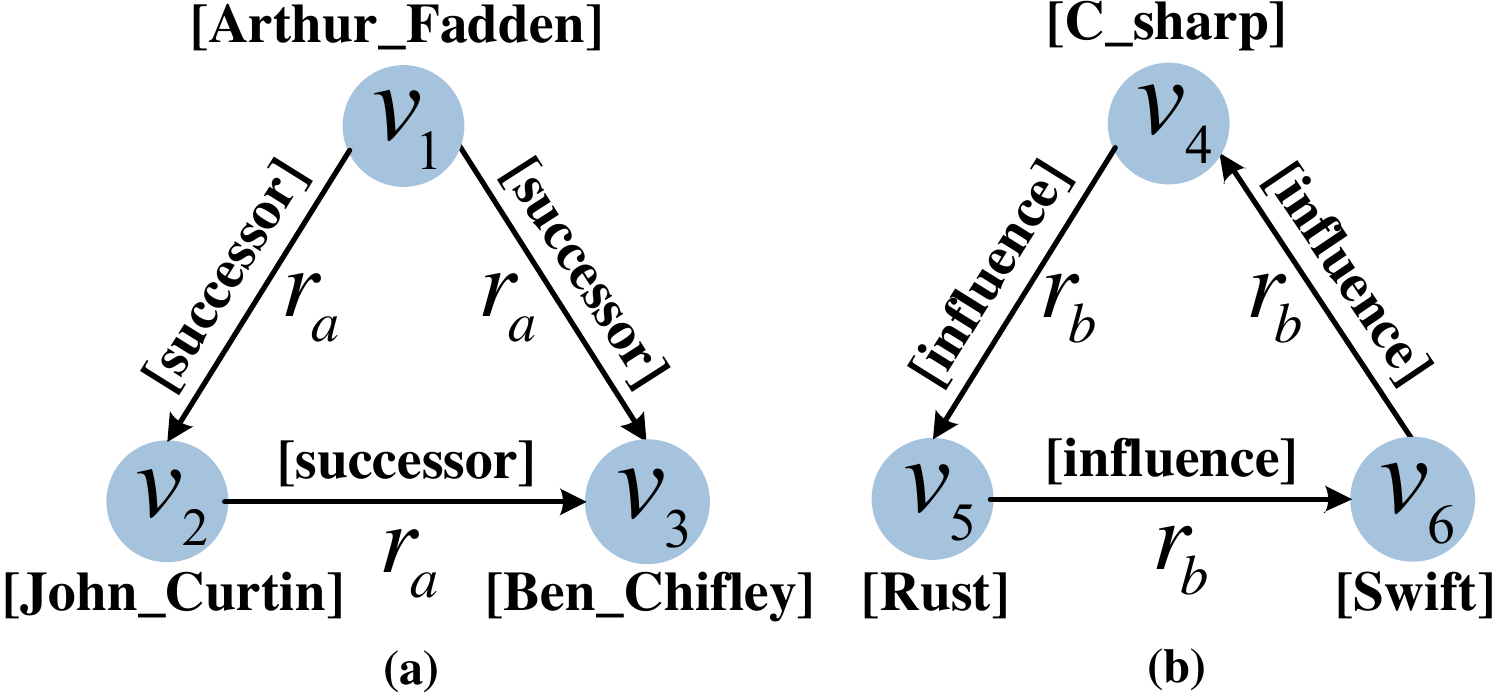}
	\caption{Examples of triangular structures (reproduced from \protect\cite{li2018structural}).}
	\label{triangular}
\end{figure}

However, KGs are usually incomplete, and different KGs are often complementary to each other. This makes a compelling case to design a technique that can integrate heterogeneous knowledge among different KGs. An effective way for doing this is \textbf{Entity Alignment}. 
There have been existing efforts devoted to embed different KGs towards entity alignment. Most of them, like JE \cite{hao2016joint}, MTransE \cite{chen2016multilingual}, JAPE \cite{sun2017cross}, IPTransE \cite{zhu2017iterative} and BootEA \cite{sun2018bootstrapping}, rely on \textit{trans-family} models to learn entity representations according to a set of prior alignments.
The most recent work \cite{wang2018cross}, takes a different approach by utilizing the Graph Convolutional Networks (GCNs) \cite{Kipf2016Semi} to jointly represent multiple KG entities, showing a new, promising direction for entity alignment. 

Compared with conventional feature based methods \cite{crowd2012,mahdisoltani2013yago3}, embedding-based methods have the advantage of requiring less human
involvement in feature construction and can be scaled to large KGs. However, there are still several hurdles that prevent wider adoption of embedding-based approaches. First, as mentioned
above, most existing methods use \textit{trans-family} models as the backbone to embed KGs, which are constrained by the assumption
$head+relation \approx tail$. This strong assumption makes it inefficient for the model to capture more complex relation information in multi-relational graphs. 

As a motivation example, Figure \ref{triangular} shows a real-world example from the \textit{DBP15K}$_{ZH-EN}$ \cite{sun2017cross} dataset. 
Prior study \cite{li2018structural} shows that trans-family methods cannot capture the triangular structures depicted in the diagram. 
For instance, for the structure of
Figure \ref{triangular}(a), TransE requires the three formulas $v_1 + r_a \approx v_2$, $v_2 + r_a \approx v_3$ and $v_1 + r_a \approx v_3$ to
hold at the same time. However, to satisfy the former two equations, we would have $v_1 + 2r_a \approx v_3$, which is contradictory to the third
equation $v_1 + r_a \approx v_3$. Accordingly, the alignment performance will inevitably be compromised if the KG representations are
learned with the trans-family, since more complex structures such as triangular ones frequently appear in multi-relational graphs.

The GCN-based model \cite{wang2018cross} represents a leap forward for embedding-based entity alignment. However, this approach is also unable to properly model relation information. Since the vanilla GCN operates on the undirected and unlabeled graphs, a GCN-based model would ignore the useful relation information of KGs. Although the Relational Graph Convolutional Networks (R-GCNs) \cite{schlichtkrull2018modeling} could be used to model multi-relational graphs, an R-GCN simply employs one weight matrix for each relation and would require an excessive set of parameters for real-world KGs that often contain thousands of relations. This drawback makes it difficult to learn an effective R-GCN model. Dual-Primal Graph CNN (DPGCNN) \cite{monti2018dual} offers a new solution for the problem. DPGCNN alternates convolution operations on the graph and its dual graph, whose vertices correspond to the edges of the original graph, and iteratively applies a graph attention mechanism to enhance primal edge representations using its dual graph. Compared with GCNs and R-GCNs, the DPGCNN can better explore complex edge structures and produce better KG representations.

Inspired by the DPGCNN, in this paper, we propose a novel Relation-aware Dual-Graph Convolutional Network (RDGCN) to tackle the challenge of proper capturing and integration for relation information. While the DPGCNN serves a good starting point, applying it to learn KG representations is not trivial. Doing so requires us to find a way to better approximate relation representations and characterize the relationship between different KG relations. We address this by extending the DPGCNN to develop a weighted model, and explore the head/tail representations initialized with entity names as a proxy to capture relation information without excessive model parameters that are often hard to train. 

As a departure from GCNs and R-GCNs, our RDGCN allows multiple rounds of interactions between the primal entity graph and its dual relation graph, enabling the
model to effectively incorporate more complex relation information into entity representations. To further integrate neighboring structural information, we also extend GCNs with highway gates. 

We evaluate our RDGCN on three real-world datasets. Experimental results show that RDGCN can effectively address the challenges mentioned above and significantly outperforms 6 recently proposed approaches on all datasets. The key contribution of this work is a novel DPGCNN-based model for learning robust KG representations. Our work is the first to extend DPGCNN for entity alignment, which yields significantly better performance over the state-of-the-art alternatives.

\section{Related Work}
\subsection{Graph Convolutional Networks}
Recently, there has been an increasing interest in extending neural networks to deal with graphs. There have been many encouraging works which are often categorized as spectral approaches \cite{bruna2014spectral,cnn_graph,Kipf2016Semi} and spatial approaches \cite{atwood2016diffusion,hamilton2017inductive,velickovic2018graph}. The GCNs \cite{Kipf2016Semi} have recently emerged as a powerful deep learning-based approach for many NLP tasks like semantic role labeling \cite{Marcheggiani2017Encoding} and neural machine translation \cite{Bastings2017Graph}. Furthermore, as an extension of GCNs, the R-GCNs \cite{schlichtkrull2018modeling} have recently been proposed to model relational data and have been successfully exploited in link prediction and entity classification. Recently, the graph attention networks (GATs) \cite{velickovic2018graph} have been proposed and achieved state-of-the-art performance. The DPGCNN \cite{monti2018dual} discussed in Section \ref{section:intro} generalizes GAT model and achieves better performance on vertex classification, link prediction, and graph-guided matrix completion tasks.

Inspired by the capability of DPGCNN on determining neighborhood-aware edge features, we propose the first relation-aware multi-graph
learning framework for entity alignment.

\subsection{Entity Alignment} 
Previous approaches of entity alignment usually require intensive expert participation \cite{crowd2012} to design model features \cite{mahdisoltani2013yago3} or an external source contributed by other users \cite{Wang2017}.
Recently, embedding-based
methods \cite{hao2016joint,chen2016multilingual,sun2017cross,zhu2017iterative,sun2018bootstrapping,wang2018cross} have been proposed to address
this issue. In addition, NTAM \cite{conf/ijcai/LiLYWSO18} is a non-translational approach that utilizes a probabilistic model for the alignment
task. KDCoE \cite{chen2018co} is a semi-supervised learning approach for co-training multilingual KG embeddings and the embeddings of entity descriptions.

As a departure from prior work, our approach directly models the relation information by constructing the dual relation graph. As we will
show later in the paper, doing so improves the learned entity embeddings which in turn lead to more accurate alignment.

\section{Problem Formulation\label{sec:problem}} 
Formally, a KG is represented as $G = (E, R, T)$, where $E, R, T$ are the sets of entities, relations and triples, respectively. Let $G_1 = (E_1, R_1, T_1)$ and $G_2 = (E_2, R_2, T_2)$ be two heterogeneous KGs to be aligned. That is, an entity in $G_1$ may have its counterpart in $G_2$ in a different language or in different surface names. 
As a starting point, we can collect a small number of equivalent entity pairs between $G_1$ and $G_2$ as the \emph{alignment seeds} $\mathbb{L} = \{(e_{i_1}, e_{i_2}) | e_{i_1} \in E_1, e_{i_2} \in E_2\}$. We define the entity alignment task as automatically finding more equivalent entities using the alignment seeds. Those known aligned entity pairs can be used as training data. 

\section{Our Approach: RDGCN}
\begin{figure*}
	\centering
	\includegraphics[width=0.9\linewidth]{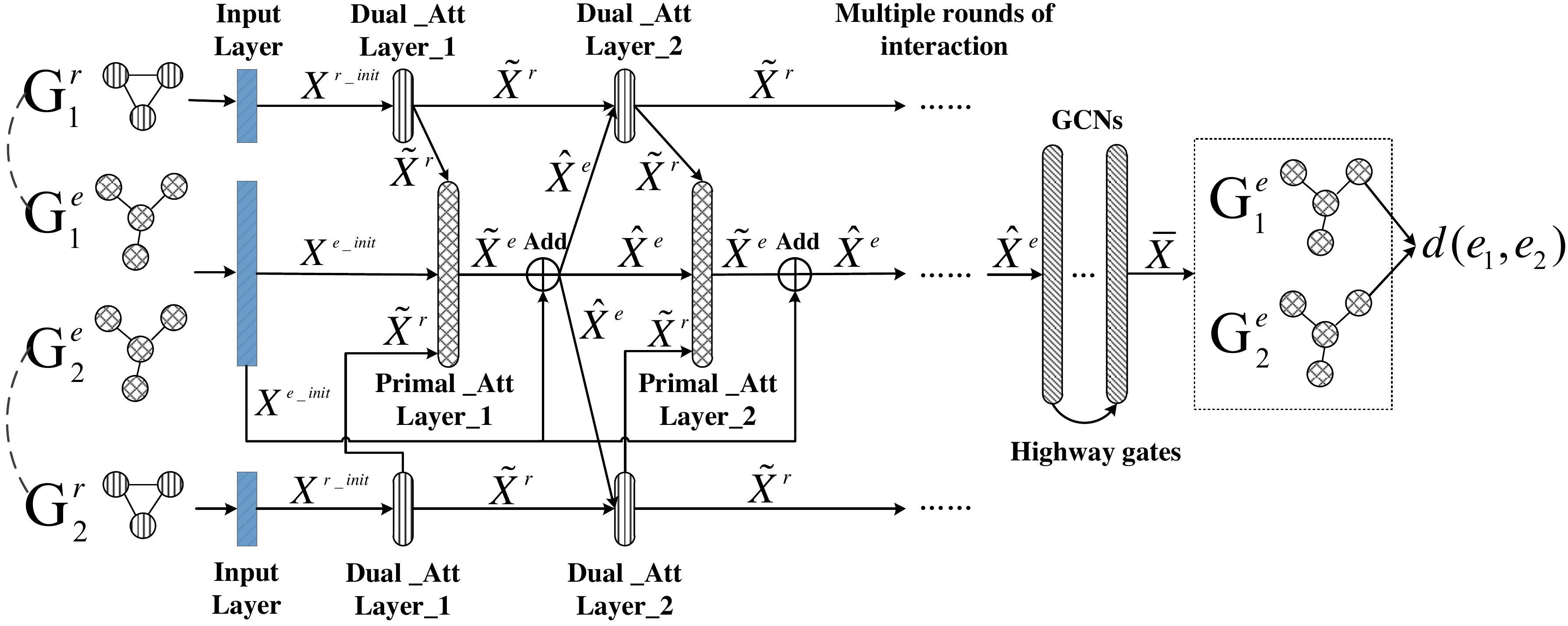}
	\caption{Overall architecture of our RDGCN. $G^r_1$ and $G^r_2$ are the dual relation graphs of $G^e_1$ and $G^e_2$, respectively. In our RDGCN model, $\mathcal{G}^e$ consists of $G^e_1$ and $G^e_2$, and $\mathcal{G}^r$ consists of $G^r_1$ and $G^r_2$.}
	\label{RDGCN}	
	\vspace{-2mm}
\end{figure*}
In order to better incorporate relation information to the entity representations, given the input KG (i.e., the primal graph), we first 
construct its dual relation graph whose vertices denote the relations 
in the original primal graph, and then, we utilize a graph attention mechanism to encourage 
interactions between the dual relation graph and the primal graph. The resulting vertex representations in primal graph are then fed to GCN \cite{Kipf2016Semi} 
layers with highway gates to capture the neighboring structural information. 
The final entity representations will be used to determine whether two entities should be aligned. Figure \ref{RDGCN} provides an overview architecture of our model.

\subsection{Constructing the Dual Relation Graph}
Without loss of generality, we put $G_1$ and $G_2$ together as the \emph{primal graph} $\mathcal{G}^e=(\mathcal{V}^e,\mathcal{E}^e)$, where the vertex set $\mathcal{V}^e = E_1 \cup E_2$ is the union of all entities in $G_1$ and $G_2$, and the edge set $\mathcal{E}^e = T_1 \cup T_2$ is the union of all edges/triples in $G_1$ and $G_2$. Note that we do not connect the alignment seeds in $\mathcal{G}^e$, thus $G_1$ and $G_2$ are disconnected in $\mathcal{G}^e$.

Given the primal graph $\mathcal{G}^e$, its \emph{dual relation graph} $\mathcal{G}^r=(\mathcal{V}^r,\mathcal{E}^r)$ is constructed as follows: 
1) for each type of relation $r$ in $\mathcal{G}^e$, there will be a vertex $v^r$ in $\mathcal{V}^r$, thus $\mathcal{V}^r=R_1 \cup R_2$; 2) if two relations, $r_i$ and $r_j$, share the same head or tail entities in $\mathcal{G}^e$, then we create an edge $u^r_{ij}$ in $\mathcal{G}^r$ connecting $v^r_i$ and $v^r_j$. 

Different from the original design of dual graph, here we expect the dual relation graph to be more expressive about the relationship between different $v^r$s in $\mathcal{G}^r$. We thus weight each edge $u^r_{ij}$ in $\mathcal{G}^r$ with a weight $w^r_{ij}$ according to how likely the two relations $v^r_i$ and $v^r_j$ share similar heads or tails in $\mathcal{G}^e$, computed as:
\begin{equation}
	\label{edge_weight}
	w^r_{ij} = H(r_i, r_j) + T(r_i, r_j)
\end{equation}
\begin{equation}
	H(r_i,r_j)=\frac{H_i\cap H_j}{H_i\cup H_j }, \mbox{  } T(r_i,r_j)=\frac{T_i\cap T_j}{T_i\cup T_j }
\end{equation}
where $H_i$ and $T_i$ are the sets of head and tail entities for relation $r_i$ in $\mathcal{G}^e$ respectively. Here, the overhead for constructing the dual graph is proportional to the number of relation types in the primal graph. In our cases, it takes less than two minutes to construct the graphs for each evaluation dataset. 

\subsection{Interactions between Dual and Primal Graphs}
Our goal of introducing dual relation graph is to better incorporate relation information into the primal graph representations. To this end, we propose to apply a graph attention mechanism (GAT) to obtain vertex representations for the dual relation graph and the primal graph iteratively, where the attention mechanism helps to prompt interactions between the two graphs. Each dual-primal interaction contains two layers, the dual attention layer and the primal attention layer. Note that we can stack multiple interactions for mutual improvement on both graphs. 

\subsubsection{Dual Attention Layer}
Let $\textbf{X}^r \in \mathbb{R}^{m \times 2d}$ denote the input dual vertex representation matrix, where each row corresponds to a vertex in the dual relation graph $\mathcal{G}^r$. Different from the vanilla GAT \cite{velickovic2018graph}, we compute the dual attention scores using the primal vertex features $\hat{\textbf{X}}^e$ (computed by Eq. \ref{primal_final}) produced by the primal attention layer from previous interaction module:

\begin{equation}
	\label{dual-update}
	\tilde{\textbf{x}}^r_i =\sigma^r(\sum_{j \in N^r_i}\alpha^r_{ij} \textbf{x}^r_j),
\end{equation}
\begin{equation}
	\alpha^r_{ij}=\frac{exp(\eta(w^r_{ij}a^r[\textbf{c}_i\|\textbf{c}_j]))}{\sum_{k \in N^r_i}exp(\eta(w^r_{ik}a^r[\textbf{c}_i\|\textbf{c}_k]))},
\end{equation}
where $\tilde{\textbf{x}}^r_i$ denotes the $d'$-dimensional output representation at dual vertex $v^r_i$ (corresponding to relation $r_i \in \mathcal{G}^e$); $\textbf{x}^r_j$ denotes the dual representation of vertex $v^r_j$; $N^r_i$ is the set of neighbor indices of $v^r_i$; $\alpha^r_{ij}$ is the dual attention score;  
$a^r$ is a fully connected layer mapping the $2d'$-dimensional input into a scalar; 
$\sigma^r$ is the activation function, ReLU; $\eta$ is the Leaky ReLU; $\|$ is the concatenation operation; $\textbf{c}_i$ is the relation representation for relation $r_i$ in $\mathcal{G}^e$ obtained from the previous primal attention layer.

Note that within our graph embedding based framework, we are not able to provide relation representations directly, due to limited training data. We thus approximate the relation representation for $r_i$ by concatenating its averaged head and tail entity representations in $\mathcal{G}^e$ as:
\begin{equation}
	\label{dualvertex}
	\textbf{c}_i=[\frac{\sum_{k \in H_i} \hat{\textbf{x}}^e_k}{|H_i|} \| \frac{\sum_{l \in T_i} \hat{\textbf{x}}^e_l}{|T_i|}],
\end{equation}
where 
$\hat{\textbf{x}}^e_k$ and $\hat{\textbf{x}}^e_l$ are the output representations
of the $k$-th head entity and $l$-th tail entity of relation $r_i$ from the previous primal attention layer.

A special case is when the current dual attention layer is the first layer of our model, we do not have $\textbf{x}^r_j$ in Eq. \ref{dual-update} produced by the previous dual attention layer, therefore, use an initial dual vertex representation produced by Eq. \ref{dualvertex} with the initial primal vertex representations $\textbf{X}^{e\_init}$. Similarly, $\textbf{c}_i$ will be obtained with the initial primal $\textbf{X}^{e\_init}$ as well. 

\subsubsection{Primal Attention Layer}
In this layer, when applying GAT on the primal graph, we can compute the primal attention scores using the dual vertex representations in $\mathcal{G}^r$, which actually correspond to the relations in the primal graph $\mathcal{G}^e$. In this way, we are able to influence the primal vertex embeddings using the relation representations produced by the dual attention layer. 

Specifically, we use $\textbf{X}^e \in \mathbb{R}^{n \times d}$ to denote the input primal vertex representation matrix. For an entity $e_q$ in primal graph $\mathcal{G}^e$, its representation $\tilde{\textbf{x}}^e_{q}$ can be computed by:

\begin{equation}
	\tilde{\textbf{x}}^e_q=\sigma^e(\sum_{t \in N^e_q}\alpha^e_{qt}\textbf{x}^e_t),
\end{equation}
\begin{equation}
	\alpha^e_{qt}=\frac{exp(\eta(a^e(\tilde{\textbf{x}}^r_{qt})))}{\sum_{k \in N^e_q}exp(\eta(a^e(\tilde{\textbf{x}}^r_{qk})))},
\end{equation}
where $\tilde{\textbf{x}}^r_{qt}$ denotes the dual representation for $r_{qt}$ (the relation between entity $e_q$ and $e_t$) obtained from $\mathcal{G}^r$; 
$\alpha^e_{qt}$ is the primal attention score; 
$N^e_q$ is the set of neighbor indices of entity $e_q$ in $\mathcal{G}^e$; 
$a^e$ is a fully connected layer mapping the $d'$-dimensional input into a scalar and $\sigma^e$ is the primal layer activation function.

In our model, the initial representation matrix for the primal vertices, $\textbf{X}^{e\_init}$, can be initialized using entity names, which provide important evidence for entity alignment. We therefore preserve the evidence explicitly by mixing the initial representations with the output of primal attention layer:
\vspace{-1mm}
\begin{equation}
	\label{primal_final}
	\hat{\textbf{x}}^e_q=\beta_s*\tilde{\textbf{x}}^e_q+\textbf{x}^{e\_init}_q,
\end{equation}
where $\hat{\textbf{x}}^e_q$ denotes the final output representation of the interaction module for entity $e_q$ in $\mathcal{G}^e$; $\beta_s$ is a weighting parameter for the $s$-th primal attention layer.

\subsection{Incorporating Structural Information}
After multiple rounds of interaction between the dual relation graph and the primal graph, 
we are able to collect relation-aware entity representations from the primal graph. Next, %
we apply two-layer GCNs \cite{Kipf2016Semi} with highway gates to the resulting primal graph to further incorporating evidence from their neighboring structures. 

In each GCN layer $l$ with entity representations $X^{(l)}$ as input, the output representations $X^{(l+1)}$ can be computed as:
\begin{equation}
	X^{(l+1)} = \xi(\tilde{D}^{- \frac{1}{2}}\tilde{A}\tilde{D}^{- \frac{1}{2}}X^{(l)}W^{(l)}),
\end{equation}
where $\tilde{A}=A+I$ is the adjacency matrix of the primal graph $\mathcal{G}^e$ with added self-connections and $I$ is an identity matrix; $\tilde{D}_{jj}=\sum_k\tilde{A}_{jk}$ and $W^{(l)} \in \mathbb{R}^{d^{(l)} \times d^{(l+1)}}$ is a layer-specific trainable weight matrix; $\xi$ is the activation function ReLU. We treat $\mathcal{G}^e$ as an undirected graph when constructing $A$, in order to allow the information to flow in both directions.

In addition, to control the noise accumulated across layers and preserve useful relation information learned from interactions, following the method described in \cite{Rahimi2018Semi}, we introduce layer-wise gates between GCN layers, which is similar in spirit to the highway networks \cite{Srivastava2015Highway}:
\begin{equation}
	T(X^{(l)})=\sigma(X^{(l)}W_T^{(l)}+b_T^{(l)}), 
\end{equation}
\vspace{-2mm}
\begin{equation}
	X^{(l+1)}= T(X^{(l)}) \cdot X^{(l+1)}+(1-T(X^{(l)})) \cdot X^{(l)} ,
\end{equation}
where $X^{(l)}$ is the input to layer $l+1$; $\sigma$ is a sigmoid function; $\cdot$ is element-wise multiplication; $W_T^{(l)}$ 
and $b_T^{(l)}$ 
are the weight matrix and bias vector for the transform gate $T(X^{(l)})$.

\paragraph{Alignment.} With the final entity representations $\bar{X}$ collected from the output of GCN layers, entity alignment can be performed by simply measuring the distance between
two entities. Specifically, the distance, $d(e_1,e_2)$, between two entities, $e_1$ from $G_1$
and $e_2$ from $G_2$ can be calculated as:
\begin{equation}
	\label{d}
	d(e_1,e_2)=\|\bar{x}_{e_1}-\bar{x}_{e_2}\|_{L_1}.
\end{equation}

\subsection{Training\label{prediction}}
For training, we expect the distance between aligned entity pairs to be as close as possible, and the distance
between negative entity pairs to be as far as possible. We thus utilize a margin-based scoring function as the
training objective:
\begin{equation}
	L=\sum\limits_{(p,q)\in \mathbb{L}}\sum\limits_{(p',q')\in \mathbb{L'}}\mathrm{max}\{0,d(p,q)-d(p',q')+\gamma\},
\end{equation}
where $\gamma > 0$ is a margin hyper-parameter; $\mathbb{L}$ is our alignment seeds and $\mathbb{L'}$ is the set of negative instances.

Rather than random sampling, we look for challenging negative samples to train our model.
Given a positive aligned pair $(p,q)$, we choose the $\mathcal{K}$-nearest
entities of $p$ (or $q$) according to Eq. \ref{d} in the embedding space to replace $q$ (or $p$) as  the negative instances. 

\section{Experimental Setup}\label{setup}
\paragraph{Datasets.}
We evaluate our approach on three large-scale cross-lingual datasets from DBP15K \cite{sun2017cross}. These
datasets are built upon Chinese, English, Japanese and French versions of DBpedia. Each dataset contains data from two KGs in different
languages and provides 15K pre-aligned entity pairs. Table \ref{dataset} gives the statistics of the datasets.
We use the same training/testing split with previous works \cite{sun2018bootstrapping}, 30\% for training and 70\% for testing. Our source code and datasets are freely available online \footnote{https://github.com/StephanieWyt/RDGCN}.

\begin{table}[t!]
	\centering
	\scriptsize
	\begin{tabular}{l|l|ccc}
		\toprule
		\multicolumn{2}{c|}{\textbf{Datasets}} & \textbf{Entities} & \textbf{Relations} & \textbf{Rel. triples} \\
		\midrule
		\multirow{2}{*}{$DBP15K_{ZH-EN}$} & Chinese & 66,469 & 2,830 & 153,929 \\
		& English & 98,125 & 2,317 & 237,674  \\
		\midrule
		\multirow{2}{*}{$DBP15K_{JA-EN}$} & Japanese & 65,744 & 2,043 & 164,373  \\
		& English & 95,680 & 2,096 & 233,319 \\
		\midrule
		\multirow{2}{*}{$DBP15K_{FR-EN}$} & French & 66,858 & 1,379 & 192,191  \\
		& English & 105,889 & 2,209 & 278,590 \\
		\bottomrule
	\end{tabular}
	\caption{Summary of the DBP15K datasets.}
	\label{dataset}
	\vspace{-1mm}
\end{table}

\paragraph{Comparison models.} We compare our approach against 6 more recent alignment methods that we have mentioned in Section \ref{section:intro}:
JE \cite{hao2016joint}, MTransE \cite{chen2016multilingual}, JAPE \cite{sun2017cross}, IPTransE \cite{zhu2017iterative},
BootEA \cite{sun2018bootstrapping} and GCN \cite{wang2018cross}, where the BootEA achieves the best performance on DBP15K.

\paragraph{Model variants.} To evaluate different components of our model, we provide four implementation
variants of RDGCN for ablation studies, including (1) GCN-s: a two-layered GCN with entity name 
initialization but no highway gates; (2) R-GCN-s: a two-layered R-GCN \cite{schlichtkrull2018modeling} with entity name
initialization; (3) HGCN-s: a two-layered GCN with entity name initialization and highway gates; (4) RD: an implementation of two dual-primal interaction modules, but without the subsequent GCN layers.

\paragraph{Implementation details.}
The configuration we used is: $\beta_1=0.1$, $\beta_2=0.3$, and $\gamma=1.0$. The dimensions of hidden representations in dual and primal attention layers are $d=300$, $d'=600$, and $\tilde d=300$. All dimensions of hidden representations in GCN layers are 300. The learning rate is set to 0.001 and we sample $\mathcal{K}=125$ negative pairs every 10 epochs.
In order to utilize entity names in different KGs for better initialization, we use Google Translate to translate Chinese,
Japanese, and French entity names into English, and then use pre-trained English word vectors
\emph{glove.840B.300d} \footnote{http://nlp.stanford.edu/projects/glove/} to construct the input entity representations for the primal graph. Note that Google Translate can not guarantee accurate translations for named entities without any context. We manually check 100 English translations for Japanese/Chinese entity names, and find around 20\% of English translations as incorrect,
posing further challenges for our model. 

\paragraph{Metrics.} We use Hits@k, a widely used metric \cite{sun2018bootstrapping,wang2018cross} in
our experiments. A Hits@k score (higher is better) is computed by measuring the proportion of correctly aligned entities ranked in the top $k$ list.

\section{Results and Discussion\label{sec:results}}

\begin{table}[t!]
	\centering
	\small
	\begin{tabular}{l|cc|cc|cc}
		\toprule
		\multirow{2}{*}{\bf Models} & \multicolumn{2}{c|}{ZH-EN} & \multicolumn{2}{c|}{JA-EN} & \multicolumn{2}{c}{ FR-EN}  \\
		& \tiny Hits@1 & \tiny Hits@10 & \tiny Hits@1 &\tiny Hits@10 & \tiny Hits@1 &\tiny Hits@10\\
		\midrule
		JE & 21.27 & 42.77 & 18.92 & 39.97 & 15.38 & 38.84 \\
		MTransE & 30.83 & 61.41 & 27.86 & 57.45 & 24.41 & 55.55 \\
		JAPE & 41.18 & 74.46 & 36.25 & 68.50 & 32.39 & 66.68 \\
		IPTransE & 40.59 & 73.47 & 36.69 & 69.26 & 33.30 & 68.54 \\
		BootEA  & 62.94 & \bf 84.75 & 62.23 & 85.39 & 65.30 & 87.44 \\
		GCN & 41.25 & 74.38 & 39.91 & 74.46 & 37.29 & 74.49 \\
		\midrule
		\bf GCN-s & 50.82 & 79.15 & 53.09 & 82.96 & 54.49 & 84.73 \\
		\bf R-GCN-s& 46.57 & 74.29 & 48.68 & 77.82 & 51.11 & 80.07 \\
		\bf HGCN-s & 69.65 & 82.53 & 75.54 & 87.87 & 88.09 & 95.27 \\
		\bf RD & 61.81 & 73.83 & 68.54 & 80.22 & 84.64 & 91.98 \\
		\bf RDGCN & \bf 70.75 &  84.55 & \bf 76.74 & \bf 89.54 & \bf 88.64 & \bf 95.72 \\
		\bottomrule
	\end{tabular}
	\caption{The overall alignment performance for all models on the DBP15K datasets. Numbers in bold indicate the best performance.}
	\label{cross}
\end{table}

\subsection{Main Results\label{overall}}
Table \ref{cross} shows the performance of all compared approaches on the evaluation datasets. By using a bootstrapping process to
iteratively explore many unlabeled data, BootEA gives the best Hits@10 score on \textit{DBP15K}$_{ZH-EN}$ and clearly outperforms GCN and other translation-based models.
It is not surprising that GCN outperforms most translation-based models, i.e., JE, MTransE, JAPE and IPTransE.
By performing graph convolution over an entity's neighbors, GCN is able to capture more structural characteristics of knowledge graphs, especially when using more GCN layers, while the translation assumption in translation-based models focuses more on the relationship among heads, tails and relations.

We observe that RDGCN gives the best performance across all metrics and datasets, except for Hits@10 on \textit{DBP15K}$_{ZH-EN}$
where the performance of RDGCN is second to BootEA with a marginally lower score (84.55 vs 84.75).
While BootEA serves a strong baseline by showing what can be achieved by exploiting many unlabeled data, our RDGCN has the advantage of
requiring less prior alignment data to learn better representations. We believe that a bootstrapping process can further improve the
performance of RDGCN, and we leave this for future work. Later in Section \ref{sec:prioralignmentdata}, we show that RDGCN maintains
consistent performance and significantly outperforms BootEA when the training dataset size is reduced. The good performance of RDGCN is
largely attributed to its capability for learning relation-aware embeddings. 

\begin{figure*}[t!]
	\centering
	\includegraphics[width=0.88\linewidth, trim=0 0 0 0, clip]{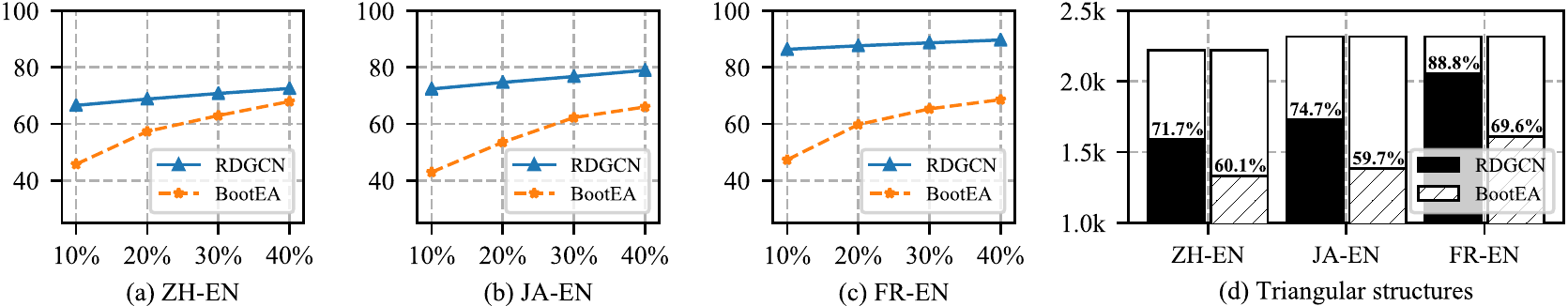}
	\caption{(a), (b) and (c) show the performance of RDGCN and BootEA using different proportions of prior entity alignments on the DBP15K datasets. The x-axes are the proportions of prior alignments, and the y-axes are Hits@1 scores. (d) shows the performance of RDGCN and BootEA on triangular structures. The x-axis is the datasets and y-axis is the number of correctly predicted pairs.}
	\label{analysis}
	\vspace{-2mm}
\end{figure*}

\subsection{Ablation Studies}

\paragraph{GCN-s vs. GCN.} As shown in Table \ref{cross}, GCN-s considerably improves GCN in all datasets, resulting in a
17.2\% increase on Hits@1 on \textit{DBP15K}$_{FR-EN}$. As mentioned in Section \ref{setup}, the three cross-lingual datasets require us to handle cross-lingual data through rough machine translations, which is likely to introduce lots of noise ($\sim$80\% accuracy in our pilot study). But our improvement over GCN shows that although noisy in nature, those rough translations can still provide useful evidence to capture, thus should not be ignored.

\paragraph{GCN-s vs. R-GCN-s.} R-GCN is an extension of GCN by explicitly modeling the KG relations, but in our experiments, we observe that GCN-s achieves better performance than R-GCN-s on all datasets. As discussed in Section \ref{section:intro}, R-GCN usually requires much more training data to learn an effective model due to its large number of parameters, and the available training data in our evaluation might not be sufficient for fully
unlocking the potential of R-GCN.

\paragraph{HGCN-s vs. GCN-s.}
Comparing HGCN-s with GCN-s, we can see that HGCN-s greatly boosts the performance of GCN-s after employing the layer-wise highway gates, e.g., over 30\% improvement of Hits@1 on
\textit{DBP15K}$_{FR-EN}$. This is mainly due to their capability of preventing noisy vertices from driving the KG representations. 

\paragraph{HGCN-s vs. RDGCN.} When comparing HGCN-s with RDGCN, we can see that the dual-primal interaction modules are crucial to the performance: removing the dual and primal attention layers leads to a drop of
1.1\% on Hits@1 and 2.02\% on Hits@10 on \textit{DBP15K}$_{ZH-EN}$. The interaction modules explore
the relation characteristics of KGs by introducing the approximate relation information and fully integrate the relation and entity information after
multiple interactions between the dual relation graph and the primal graph. The results show that effective modeling and use of relation information is
beneficial for entity alignment.

\paragraph{RD vs. RDGCN.} Comparing RD with RDGCN, there is a significant drop
in performance when removing the GCN layers from our model, e.g., the Hits@1 of RD and RDGCN
differ by 8.94\% on \textit{DBP15K}$_{ZH-EN}$. This is not surprising, because the dual-primal graph interactions are designed to integrate KG relation information, while the GCN layers can effectively capture the neighboring structural information of KGs. These two key components are, to some extent, complementary to each other, and should be combined together to learn better relation-aware representations.

\subsection{Analysis\label{sec:prioralignmentdata}} 

\paragraph{Triangular structures.}
Figure \ref{analysis}(d) shows the performance of RDGCN and BootEA, the state-of-the-art alignment model, on the testing instances with triangular structures. 
We can see that the alignment accuracy of our RDGCN for entities with triangular structures is significantly higher than that of BootEA in all three datasets, showing that RDGCN can better deal with the complex relation information. 

\paragraph{Impact of available prior alignments.}
We further compare our RDGCN with BootEA by varying the
proportion of pre-aligned entities from 10\% to 40\% with a step of 10\%. As expected, the results of both models on all three datasets
gradually improve with an increased amount of prior alignment information. According to Figure \ref{analysis}(a-c), our RDGCN consistently outperforms BootEA, and seems to be insensitive to the proportion of prior alignments. When only using 10\% of the pre-aligned entity pairs as training data, RDGCN still
achieves promising results. For example, RDGCN using 10\% of prior alignments achieves 86.35\% for Hits@1 on \textit{DBP15K}$_{FR-EN}$. This result
translates to a 17.79\% higher Hits@1 score over BootEA when BootEA uses 40\% of prior alignments. 
These results further confirm the robustness of our model, especially with limited prior alignments. 

\begin{figure}[t!]
	\centering
	\includegraphics[width=0.9\linewidth]{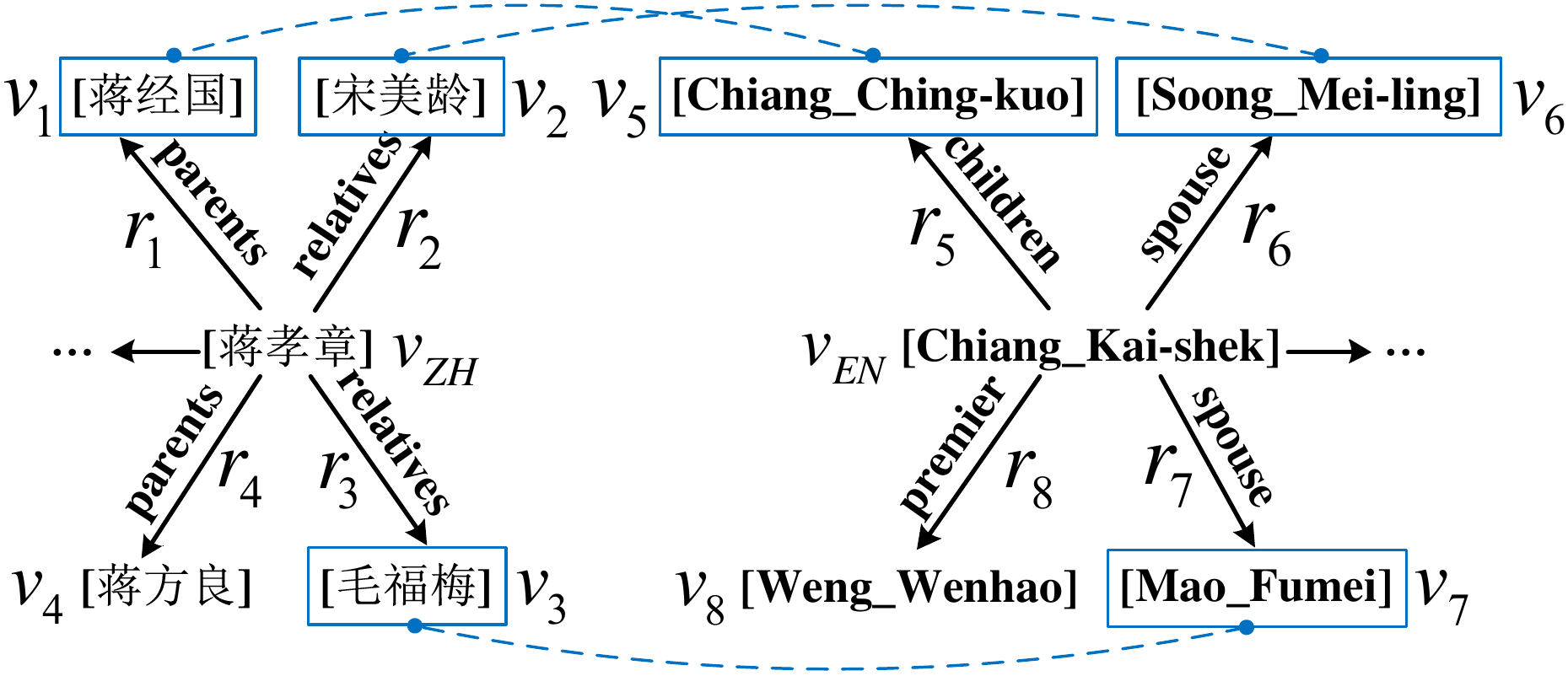}
	\caption{An example in \textit{DBP15K}$_{ZH-EN}$, where the blue dash lines indicate the connected entities should be aligned.
	}
	\label{casestudy}
\end{figure}

\paragraph{Case study.}
Figure \ref{casestudy} shows an example in \textit{DBP15K}$_{ZH-EN}$ 
and the target entity pair, $(v_{ZH}$ and $v_{EN})$, should not be aligned. 
The competitive translation-based models, including BootEA, give lower distance scores for $(v_{ZH}$ and $v_{EN})$, suggesting that these two entities should be aligned. This is because those models fail to address the specific relation information associated with the three aligned neighboring entities. For this example, both $v_1$ and $v_5$ indicate the person \emph{Chiang\_Ching-kuo}, but $v_1$ has the relation \emph{parents} with $v_{ZH}$, while $v_2$ has the relation \emph{children} with $v_{EN}$.
Utilizing such information, a better alignment model should produce a
larger distance score for the two entities despite they have similar neighbors. By carefully considering the relation information during
the dual-primal interactions, our RDGCN gives a larger distance score, leading to the correct alignment result.

\section{Conclusions}    

This paper presents a novel Relation-aware Dual-Graph Convolutional Network for entity alignment over heterogeneous KGs. Our approach is designed to explore complex relation information that commonly exists in multi-relational KGs. 
By modeling the attentive interactions between the primal graph and dual relation graph, our model is able to incorporate relation information with neighboring structural information through gated GCN layers, and learn better entity representations for alignment. Compared to the state-of-the-art methods, our model uses less training data but achieves the best alignment performance across three real-world datasets.

\section*{Acknowledgements}
This work is supported in part by the NSFC (Grant No. 61672057, 61672058, 61872294), the National Hi-Tech R\&D Program of China (No. 2018YFC0831905), and a UK Royal Society International Collaboration Grant (IE161012). For any correspondence, please contact Yansong Feng.

\bibliographystyle{named}
\bibliography{ijcai19}

\end{document}